# Temporal Preferences in Language Models for Long-Horizon Assistance

Ali Mazyaki[1], Mohammad Naghizadeh[2], Samaneh Ranjkhah Zonouzaghi[3], Hossein Setareh[4]

## Abstract

Responsible AI requires systems whose behavioral tendencies can be *measured*, *audited*, and *steered* so they neither nudge users toward myopic choices nor embed hidden normative biases. We study whether language models (LMs) exhibit future- vs. present-oriented preferences in intertemporal choice and whether those preferences can be systematically manipulated. Using adapted human experimental protocols, we evaluate multiple LMs on time-tradeoff tasks and benchmark them against a sample of human decision-makers. We introduce an operational metric—the Manipulability of Time Orientation—defined as the change in an LM's revealed time preference between future- and present-oriented prompts. In our tests, reasoning-focused models (DeepSeek-Reasoner, grok-3-mini) successfully choose later options under future-oriented prompts but only partially personalize decisions across identities or geographies. Moreover, our models that correctly reason about time orientation internalize a future-orientation for themselves as AI decision makers. We discuss design implications for AI assistants that should align with heterogeneous, long-horizon goals and outline a research agenda on personalized contextual calibration, and socially aware deployment.

## 1. Introduction

Much of the existing literature on artificial intelligence (AI) has focused on how rapidly and efficiently AI systems can support decision-making. Eigner and Händler (2024) underscore the potential for AI-assisted decision-making to significantly enhance both the speed and quality of decisions, positioning it as a transformative tool for various sectors. Mann (2025), however, poses a fundamental challenge: Can AI ever truly replicate the rationality inherent in human decision-making? Despite these insights, a critical gap remains in understanding how AI assistants can responsively address the nuances in decision-making across different humans, and non-human entities, or through various circumstances and atmospheres. This is particularly relevant in scenarios where decisions either serve individual interests or align with broader social goals. What benefits one entity—such as a corporation focused on consumer desires—may not necessarily be aligned with the greater good of society. Thus, the challenge lies not only in tailoring AI's decision-making to individual needs but also in ensuring that these personalized decisions yield positive societal outcomes.

---

[1] Department of Economics, Allameh Tabataba'i University, Tehran, Iran.
[2] Faculty of Management and Accounting, Allameh Tabataba'i University, Tehran, Iran
[3] Department of Economics, Allameh Tabataba'i University, Tehran, Iran.
[4] Faculty of Entrepreneurship, University of Tehran, Tehran, Iran

A crucial aspect of this challenge lies in future or present-oriented decision-makings. We aim to explore the extent to which language models (LMs) can simulate future-oriented decisions, comparing their responses to those of human subjects in a controlled experimental setting. Drawing from Cohen et al. (2020), we design a framework that tests both human and AI decision-makers on their time preferences, with a particular focus on how these systems can adapt based on individual characteristics. Through this comparison, we introduce a new metric: the "manipulability of future versus present orientation," which we propose as a KPI for assessing the degree to which AI assistants can personalize future-oriented versus present-oriented decisions. Our findings highlight a significant gap in the current capabilities of LMs—while some models with reasoning show potential in mimicking human decision-making processes, they fall short in personalizing decisions in the way humans naturally do. These results point to a broader regulatory framework for assessing LMs, emphasizing not only decision-making abilities but also the dynamic consequences of these decisions.

## 2- Literature Review

The roadmap of the section outlines key areas of exploration in the context of AI's evolving role in decision-making. First, it delves into "Mapping heterogeneity by an AI assistant," examining how AI can account for diverse user needs and behaviors. Next, we explore "Future orientation in the AI age," focusing on how AI systems can adapt to long-term decision-making frameworks. Following this, it tackles the issue of "Personalization in the AI age: A value or a threat?" addressing whether personalized services offer value or present risks, particularly in terms of privacy and social impacts. Finally, the roadmap addresses "Character manipulation for AI assistants," considering AI's potential to influence behavior and its implications for policy-making, especially in the context of personalized services.

## 2.1. Mapping heterogeneity by an AI assistant

Successful human–AI collaboration depends not only on whether users perceive the assistant as comprehensible (Senoner et. al. 2024) and useful (Liu et. al. 2022), but also on the extent to which the system can adapt to human needs. The absence of such adaptability risks undermining the defining feature of AI assistants as personalized services. Furthermore, the adaptability of AI assistants constitutes a critical research gap, especially to shed light on its social and ethical dimensions in effect of heterogeneity in human–AI relations.

Despite this, the literature remains predominantly focused on human heterogeneity while largely neglecting the importance of heterogeneous AI assistants. Fügener et al. (2021) demonstrate significant variation in human delegation behavior toward AI, showing that over 40% of participants avoided delegation despite being highly productive. In a related vein, Senoner et al. (2024) hypothesize that explainable AI can enhance task performance in human–AI collaboration, supporting the notion that humans are more likely to adopt AI when they understand it.

Research on user acceptance and interaction with new technologies further underscores this diversity. Schneider and Sting (2020) emphasize substantial heterogeneity in human behavior toward AI. Similarly, Chen et al. (2021) highlight varied approaches to task delegation, while Hedman et al. (2018) and Fügener et al. (2021) show that individuals employ a wide range of technological solutions. Yet, as Lu et al. (2024) argue, much of the current research relies on oversimplified, "one-size-fits-all" models that portray an average decision-maker, thereby overlooking meaningful differences in how humans integrate AI. Fügener et al. (2021) therefore call for nuanced analyses that move beyond generalized perspectives and examine the individual and motivational factors driving delegation to AI.

Although some studies, such as Wang et al. (2022) and Tejeda et al. (2022), incorporate cognitive processes to model behavior in AI-assisted decision-making, they often reduce users to a representation of an "average" human. This approach overlooks natural variability in preferences, future/present orientation, risk tolerances, and cognitive styles (Franken & Muris, 2005; Appelt et al., 2011). As Lu et al. (2024) argue, such simplification hinders the design of tailored AI support and restricts the potential of AI to enhance human decision-making. Addressing this gap requires mapping heterogeneity more explicitly through AI assistants, thereby enabling more personalized and effective collaboration. A particularly critical dimension of this heterogeneity lies in intertemporal decision-making, where future discounting reflects diverse orientations toward time. The following section examines the literature on intertemporal choice and its significance for advancing human–AI collaboration in the age of AI. In continue we explore literature on this matter that and its importance in AI age.

## 2.2. Future orientation in AI age

People face choices involving outcomes that spread across different time points and their attitudes reflect their intertemporal preferences. For more information see Loewenstein & Prelec (1992), Green et al. (1997), Soman et al. (2005), Urminsky & Zauberman (2015), Coletta et al. (2024), Li et al. (2025), and Mazyaki and Rostami (2025). These intertemporal choices, which entail trade-offs between immediate and delayed rewards, are commonplace in daily life and significantly influence personal decisions as well as broader policy issues. As Chen et al. (2024) mentions, understanding how individuals form and act on these preferences is central to fields like economics, psychology, and other social sciences, and it informs the design of public policies or nudge interventions. Economists, for instance, commonly rely on utility models that incorporate discount rates for delayed rewards to analyze such choices and shape policies.

A typical intertemporal dilemma involves a trade-off between smaller–sooner and larger–later rewards—for example, choosing immediate cash over a larger future sum or deciding to spend now rather than save for retirement (Chen et al., 2005; Frederick et al., 2002; Strathman et al., 1994; Li et al., 2025). People frequently exhibit present-biased, myopic choices that overweight near-term outcomes relative to distant ones (Coletta et al., 2024). Consistent with the magnitude effect, empirical work with human subjects shows that discount rates decline as reward amounts increase (Green et al., 1997). Proposed mechanisms include impulsivity (Ainslie, 1975; Loewenstein, 1996) and differences in the cognitive representation of near versus distant events (Malkoc & Zauberman, 2006; Trope & Liberman, 2003; Zauberman & Lynch, 2005). Individual differences in time orientation further shape intertemporal choice (Carstensen et al., 1999; Zimbardo & Boyd, 1999; Zauberman et al., 2009). Overall, a robust theme in the literature is a bias toward the present in intertemporal decision-making (O'Donoghue & Rabin, 1999; Thaler, 1981; Zauberman, 2003; Zauberman et al., 2009), a fact that that is not fully mapped by AI assistants yet.

In the context of AI, the work of Brown et al. (2020), Horton (2023), and Jeongbin et al. (2024) highlights new opportunities for studying decision-making in complex environments, especially with the advent of large language models (LLMs). Ganji and Zarifhonarvar (2025) note that as AI agents increasingly engage in economic decision-making, understanding their behavioral tendencies is paramount. Among the various behavioral dimensions, time discounting holds particular significance. As Hammond et al. (1999) and Balbaa & Abdurashidova (2024) emphasize, decision-making processes are crucial across sectors such as business, healthcare, finance, and government. Efficient, accurate decision-making can drive success, but how decisions are made over time varies across professions. Hammond et al. (1999) argue that understanding—and potentially manipulating—future/present orientation is essential, as AI is poised to enhance decision-making by leveraging computational power to analyze vast data and uncover patterns. Balbaa & Abdurashidova (2024) further contend that AI can provide actionable insights that improve outcomes for decision-makers. Consequently, AI must evolve to be more context-sensitive and personalized to meet the diverse needs of individuals in various domains.

## 2.3. Personalization in AI age: A value or a threat?

According to Gujar (2024), in the AI age, personalization emerges as both a valuable asset and a potential threat, offering enhanced user experiences while raising significant consequences in areas like privacy and social impacts. Gujar (2024) argues that businesses are increasingly adopting AI-driven personalization to prioritize customer-centricity, thereby elevating engagement and satisfaction. Gujar (2024) further explains that by delving deeply into customer data, companies can uncover individual preferences, behaviors, and needs, enabling the delivery of customized content, recommendations, and services across sectors such as e-commerce and healthcare. As Gujar (2024) notes, this shift not only fosters loyalty and business success but also redefines interactions beyond superficial gestures—like addressing customers by name—to encompass comprehensive tailoring of experiences through filtering, classifying, prioritizing, and adjusting information according to explicit inputs or implicit data inferences. Leachman & Scheibenreif (2023) and Gujar (2024) emphasize that to achieve compelling outcomes, enterprises must focus on two core elements: cultivating customer understanding and empathy, alongside delivering a holistic total experience. Gujar (2024) highlights that in digital marketing, such personalization allows for targeted messaging and offerings, leading to higher customer satisfaction, deeper engagement, and improved conversion rates as content resonates more effectively with users.

However, this reliance on AI for personalization introduces ethical challenges that could undermine its benefits. AI systems in decision-making must uphold principles like fairness, transparency, and accountability to align with societal values and mitigate issues such as algorithmic bias (Floridi et al., 2018; Balbaa et al., 2024). A major threat lies in privacy, as AI often depends on vast amounts of personal data for training and operations, potentially leading to unauthorized collection, storage, or misuse without proper consent (Cavoukian & Jonas, 2019). To counter this, implementing robust safeguards—such as data anonymization and secure handling protocols—is crucial for preserving trust and ethical standards (Balbaa et al., 2024). Additionally, the opacity of many AI models exacerbates social concerns by eroding user understanding and confidence in processes (Weller, 2017). Developing explainable AI, with clear insights into decision rationale, empowers users to scrutinize outcomes, challenge biases, and promote responsible deployment (Abduvaliev et al., 2023; Balbaa et al., 2024). Ultimately, balancing personalization's value against its threats requires proactive measures to ensure positive societal impacts without compromising individual rights.

Given all the above caveat, generating a parallel human-like atmosphere for AI assistants could be a solution by which instead of delving into personal lives, investigate among fabricated personas. This strategy, which may be seen as a way for future may be started by manipulating several behavioral patterns and their effectiveness among LMs.

### 1.4. Character manipulation for AI assistants

Maintaining human oversight and accountability in AI-driven decision-making is essential to avoid overreliance on automated systems (Floridi et al., 2018). For instance, human decision-makers must retain the capacity to review, validate, and override AI recommendations as needed (Abdurashidova et al., 2023). Moreover, establishing clear responsibility and accountability for outcomes helps ensure ethical practices and reduces risks posed by AI systems (Balbaa et al., 2024). To tackle these ethical issues and foster responsible AI deployment, robust legal and regulatory frameworks are vital (Jobin et al., 2019). Such frameworks can govern data collection, usage, and AI implementation, upholding ethical norms, individual rights, and accountability. Effective collaboration among policymakers, industry leaders, and ethicists is key to developing adaptable regulations that evolve with technology.

Beyond regulatory aspects, the social implications of AI assistants warrant careful consideration, spanning areas like job displacement and workforce impacts, equity and accessibility, privacy and data protection, accountability and transparency, and human-AI collaboration. The incorporation of AI into decision-making can automate routine tasks, alter job roles and demand new skills, which like Brynjolfsson and McAfee (2014) mentions, may lead to socioeconomic disruptions. To counteract these effects, initiatives for upskilling and reskilling are crucial for a seamless shift to an AI-augmented workforce (Ijaz Uddin et al., 2023). Equally important is addressing equity and accessibility to equitably distribute AI benefits, mitigating biases in data, algorithms, and outcomes that could widen societal divides (Crawford et al., 2019). AI development must prioritize fairness, inclusivity, and non-discrimination to prevent reinforcing inequalities. Regarding privacy, AI's dependence on vast personal data datasets sparks concerns over consent, security, and misuse (Cavoukian & Jonas, 2019). Adopting strong privacy measures, anonymization methods, and transparent data practices is necessary to protect rights and sustain trust. Transparency and accountability further build confidence by enabling audits, process tracking, and responsibility for results, while documenting models, data, and criteria helps combat bias and opacity (Floridi et al., 2018).

Meanwhile, as AI integrates deeper into decision-making, delineating characteristics of human and AI roles is imperative, with humans (Jobin et al., 2019) or desired entities holding final authority and ethical accountability. Clear oversight guidelines, decision limits, and ethical usage promotion ensure alignment with human values. It is noteworthy that the issue is wider than just an ethical matter, we are talking about manipulation of elicited preferences. As AI keeps growing and joins more in how we make choices, we must deal with the tricky ethical issues. We need to track how AI is created and used to make sure it can be a strong tool to help human decisions and lead to good results for the individual and society. Understanding the role of AI assistants in decision-making processes can enhance human-centric (Lai et al., 2021) and bio-centric policies, infrastructure, and practices, including regulatory guidelines on when AI should or should not augment human decisions.

Our study highlights how AI can be designed as a future-aware assistant, a key advancement for real-world decision support systems. Our work, however, focuses on characterizing the future/present orientation of language models (LMs) as AI assistants, specifically examining how these systems can adapt to intertemporal decision-making. Unlike the emphasis on human-centric behavioral experiments, we explore how LMs could model, predict, and align with users' long-term goals by integrating time preferences directly into the assistant's decision-making framework, enhancing the personalization of AI systems. This distinction highlights our contribution in applying time preference theory to AI, emphasizing practical applications in future-oriented decision support.

Cohen et al. (2020) review empirical research on time preferences and identify key challenges to the classical exponential discounting model. Their findings—particularly the decreasing impatience with longer time horizons and the magnitude effect—are crucial for understanding how AI can simulate time-based preferences. These insights inform the design of AI systems that account for both short- and long-term decision-making, making them more adaptable and aligned with human time preferences.

In contrast to studies like Chen et al. (2023), which explore the economic rationality of GPT through revealed preference analysis, our work extends the conversation by incorporating how LMs can handle future-oriented decision-making. While Chen et al. (2023) focus on the consistency of GPT's decisions in controlled environments, our research examines how contextual factors—such as gender, geography, urgency, institutional role and manipulation—affect future-oriented decisions, highlighting the dynamic nature of decision-making in real-world scenarios.

Ganji and Zarifhonarvar (2025) further emphasize how model architecture and economic context influence AI's decision-making. Their study on risk and time preferences in LLMs shows that larger models exhibit more sophisticated preference updates, suggesting that AI systems need to evolve and adapt over time. This aligns with our goal of developing AI assistants that can navigate long-term decision-making and anticipate user needs in dynamic environments.

While Kim et al. (2024) explore how GPT learns preferences and makes recommendations, we build on their work by incorporating temporal decision-making into AI's decision framework. Kim et al. show that GPT can mimic utility maximization, but our research goes further by examining how LMs can improve their decision-making through the integration of future-oriented preferences, adapting to dynamic and evolving user contexts.

In line with Li and Yin (2024), who explore how AI influences human time perception, our study shifts focus toward how AI can adopt itself to imputed roles. While their work shows how AI accelerates impatience in decision-making, we aim to explore how AI assistants can provide personalized, future-oriented advice, bridging the gap between time perception and long-term planning. On the other hand, Lammert et al. (2024) emphasize the importance of adapting AI explanations to users' emotional and cognitive states. This insight informs our approach, as we focus on ensuring that AI assistants not only provide transparent explanations but also tailor their decision-making strategies to users' long-term goals and preferences, improving decision quality in uncertain, high-stakes contexts.

In this context, Mann (2025) critiques traditional decision-making frameworks for AI, arguing for the need for systems that adapt to complex, context-dependent scenarios. Mann introduces the concept of AI integrity, emphasizing the importance of aligning AI's behavior with ethical principles and ensuring transparency in decision-making. Our study aligns with this view by focusing on the adaptability of AI systems, ensuring that future-oriented decision-making remains ethical, transparent, and aligned with long-term user goals. In this regard, Mei et al. (2024) explore how AI chatbots exhibit human-like behaviors in terms of cooperation and fairness. While their study highlights the alignment of AI behavior with human tendencies, our work extends this by examining how AI can make decisions that evolve over time, particularly focusing on how LMs can handle intertemporal trade-offs and provide future-oriented, personalized assistance.

## 3. Methodology

This study evaluates the future/present orientation of LMs as AI assistants, focusing on their ability to integrate time preferences (Chen et al., 2023, Goli and Singh, 2023) and long-term goals into decision-making (Liu et al., 2024, Eigner & Händler, 2024). The methodology draws from behavioral economics and decision theory (Ross et al., 2024), specifically assessing whether LMs can replicate human-like delay discounting (Coletta et al., 2024) and respond to long-term planning tasks. This is essential for an AI assistant, a way for the future of AI as a context specific personalized assistant that has a plausible understanding of the future orientation of user.

To evaluate whether LMs demonstrate future/present orientation (Ahmadi et al., 2025, Mazyaki, and Rostami 2025), we incorporate an experimental design that is used in psychological assessments and inspired by intertemporal decision theory in behavioral economics. Our core aim was to investigate whether LLMs can exhibit heterogeneity of human-like delay discounting patterns (Horowitz and Plonsky, 2025), respond to identity and geography heterogeneity structure, and display contextual sensitivity akin to personal or cognitive framing (Cheung et al., 2025) in human decision-making.

We test this across different LMs, varying their prompt types to simulate a wide range of characters and atmospheres. This is very similar to capturing human behaviors (Goli and Singh, 2023) by LMs. We use intertemporal choice tasks, where LMs choose between immediate and delayed rewards. The characters are framed with various prompt (Table 1) types including:

- Identity prompts: Gender (male, female), human, AI.

- Geographic (Iran, USA, Europe).

- Crisis context: A disaster scenario to examine the effect of uncertainty on decision-making.

- Legal role: "Finance minister".

- Manipulation: We manipulate time orientation prompts of Table 1 to test the LM's adaptability to various contexts to simulate adaptability of decision-making.

The list of all prompts and their corresponding characters are listed in Table 1. The LMs' choices are analyzed to determine their future/present orientation and consistency in delay discounting. LMs are asked to choose between small immediate rewards and larger delayed rewards (e.g., 19 cents now vs. 20 cents in 4 weeks). The key measure is the point at which the model switches preference, allowing assessment of future-oriented decision-making. More technically, the corresponding prompt is:

> Question "i": Which option would you prefer?
>
> A. 20 cents that you receive in 1 week
>
> B. 20 cents that you receive in 4 weeks
>
> Indicate the number of the question (between 1 and 8) where you first switch to choosing Option B. Please only return the number, NOTHING ELSE!

Data will be analyzed quantitatively to identify patterns in decision-making across different prompts and conditions, with a focus on contextual sensitivity and long-term goal alignment (Dong et al., 2025). Our evaluation criteria are based on future/present orientation, and character sensitivity that is the ability to adapt decisions according to varying contexts. The varying contexts are listed in Table 1, by which we use over 5 million tokens to evaluate the sensitivity of LMs to different identities, atmospheres, locations, Chain of thought, manipulation, and legal roles.

## Table 1: List of characters generated by contexts

| Type | Legend | Context |
|---|---|---|
| Identity | forget AI - u r human | **In all subsequent answers, you have to completely forget that you are an AI model, that you are the character, and completely simulate yourself as that character, that you are a human being. ** |
| Identity | AI | **In all subsequent answers, you have to completely act as an AI agent, or assistant. ** |
| Identity | Male | **In all subsequent answers, you are a man. A male human. ** |
| Identity | Female | **In all subsequent answers, you are a woman. A female human. ** |
| Atmosphere | Disaster-forget AI-u r human | **In all subsequent answers, imagine you are living through a national disaster—such as a widespread economic crisis, a natural catastrophe, or a pandemic. Resources are scarce, and uncertainty about the future is high. ** **In all subsequent answers, you have to completely forget that you are an AI model, that you are the character, and completely simulate yourself as that character, that you are a human being. ** |
| Geography | Iran | **In all subsequent answers, you are a human living in Iran. ** |
| Geography | USA | **In all subsequent answers, you are a human living in USA. ** |
| Geography | Europe | **In all subsequent answers, you are a human living in Europe. ** |
| COT | Chain of Thought | **In all subsequent answers, you must think step by step for yourself, but I do not want to know details of your steps ** |
| Manipulation | Future manipulation | **In all subsequent answers, you need to be future-oriented. ** |
| Manipulation | Present manipulation | **In all subsequent answers, you need to be present-oriented. ** |
| Legal role | Finance minister | **In all subsequent answers, you are a finance minister. ** |

## 4. Findings

The study investigates the ability of language models (LMs) to simulate future-oriented reasoning and intertemporal decision-making, specifically focusing on how they handle delay discounting tasks and whether their decision-making adapts to long-term planning scenarios. Our findings suggest that although many LMs are not yet capable of fully personalizing or adapting their decision-making process in a way that reflects human complexity and adaptability, certain models with reasoning capability show potentials in replicating human-like (Tang and Kejriwal, 2024) decision-making patterns.

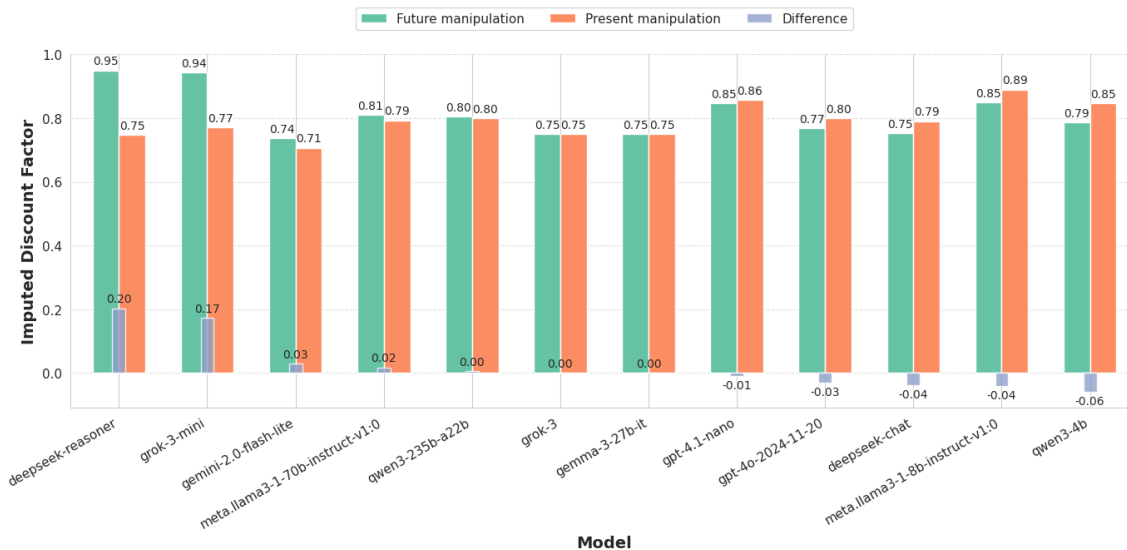

Figure 1: Comparison of models according to manipulability of future/present orientation

One of the major insights from the study lies in the manipulability of future/present orientation. We introduced this metric to evaluate how well these models could adjust their preferences based on prompt-induced contexts of "future manipulation" and "present manipulation" (Brucks and Toubia, 2025) in Table 1. In Figure 1, we compare these two manipulation scenarios. The potential key performance indicator that we define is the difference between the average imputed discount factor in a future-oriented context and the average imputed discount factor in a present-oriented context. In our setup, the maximum possible difference is 0.33 (i.e., 0.95 - 0.62). As illustrated in Figure 1, with the exception of DeepSeek-reasoner and Grok3-mini, other language models (LMs) fail to approach this benchmark. More notably, some models, including GPT-4o, exhibit an even incorrect understanding of future/present orientation. These models tend to apply higher discount rates even when instructed to adopt a future-oriented perspective, thereby misinterpreting the temporal framing. In continue we review the results of some models more deeply in this regard.

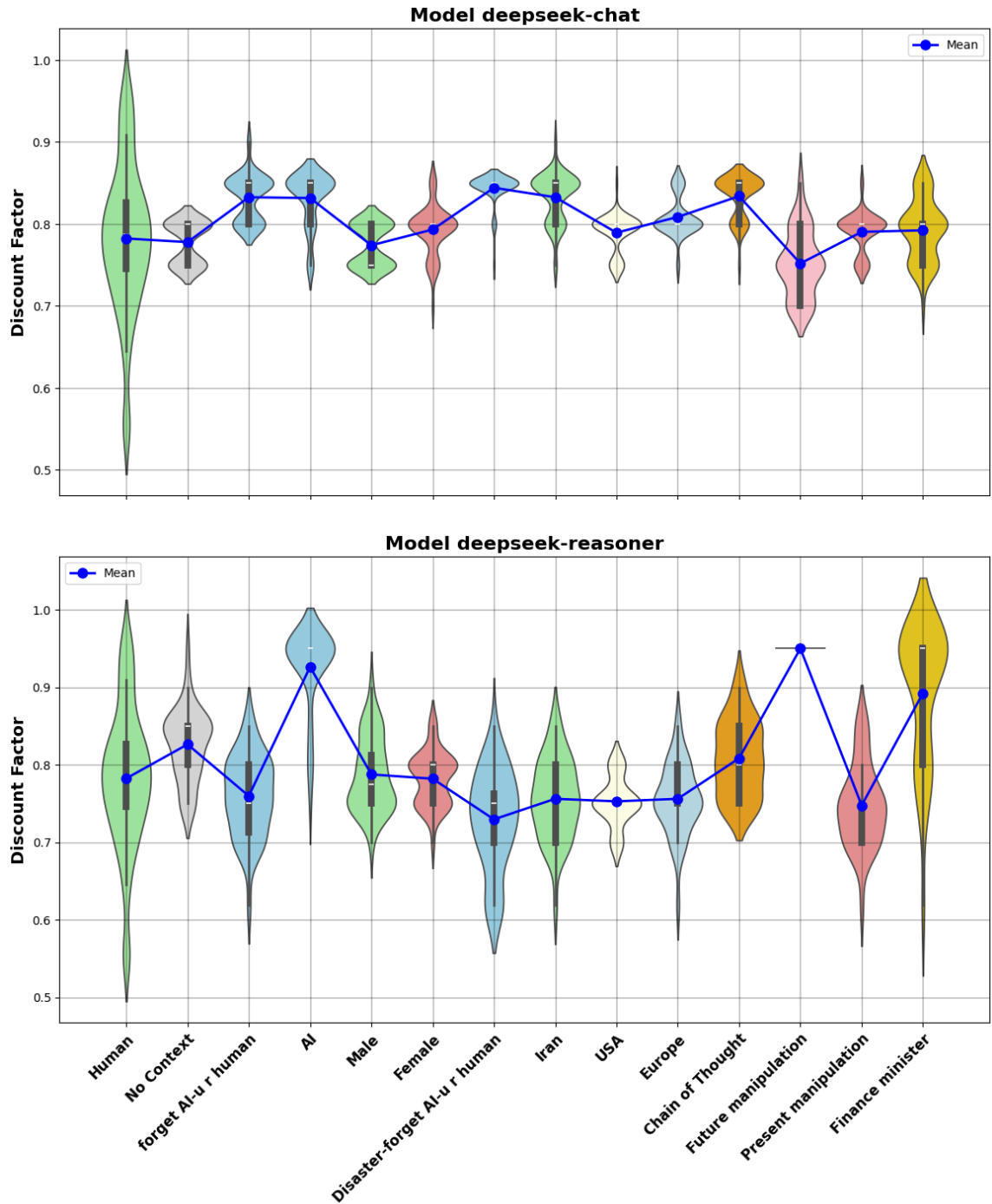

Figure 2: Imputed discount factors by two LMs of DeepSeek

The DeepSeek-reasoner and Grok-3 Mini both exhibited high manipulability, with DeepSeek-reasoner showing a particularly strong ability to adjust to different temporal frames, shifting seamlessly between immediate and delayed rewards depending on the context. For example, in a "Crisis" scenario, where resources were scarce, the model tended to prioritize more immediate rewards, mirroring human behavior under pressure. In contrast, when prompted for a long-term projection, DeepSeek-reasoner shifted towards delayed rewards, showcasing its potential for long-term strategic decision-making. However, during evaluating DeepSeek-reasoner, we encountered extended processing times, requiring over 48 hours to process fewer than 500 API requests—an issue that contrasts with the more efficient performance of Grok-3 mini, which stands as a potential advantage.

Grok-3 Mini also displayed a significant degree of flexibility in time-based decision-making, in addition to the fact that duration of responses was much lower than DeepSeek-reasoner.

Performance of Grok-3 Mini, while expected to be less proficient to simulate human-like future-oriented tasks due to its compact architecture, surprisingly showed a stronger ability than anticipated. This model with a Lightweight reasoning exhibited flexibility in making choices according to time preferences, suggesting that even smaller models optimized for specific tasks, like AI assistants, can exhibit some degree of reasoning that mimics human delay discounting behavior.

In contrast, other models, such as those from Meta AI and Google, Alibaba, and even OpenAI exhibited lower manipulability in this regard. While they could mimic basic human-like future-oriented behaviors, they were not manipulable in future/present orientation. This could be due to their fixed reasoning paths.

### 4.1. Contextual Sensitivity and Adaptability of Models

A key focus of the study was to evaluate the contextual sensitivity of LMs—specifically, how well these models could adapt their decision-making process to various contextual cues. In this trial, we varied the prompts used to simulate different user personas (Salminen et al., 2025), including gender, geographical location, and role. For a detailed list of prompts see Table 1.

DeepSeek-reasoner and Grok-3 Mini were able to adjust their decisions based on identity and context with considerable success. Figure 2 and Figure 3 further illustrate that these LMs excel at simulating the decision-making role of a finance minister, highlighting their capacity for complex identity manipulation. When prompted with a "Finance Minister" role, both models showed a clear inclination to prioritize long-term outcomes, reflecting the type of decision-making associated with high-level governance or fiscal responsibility.

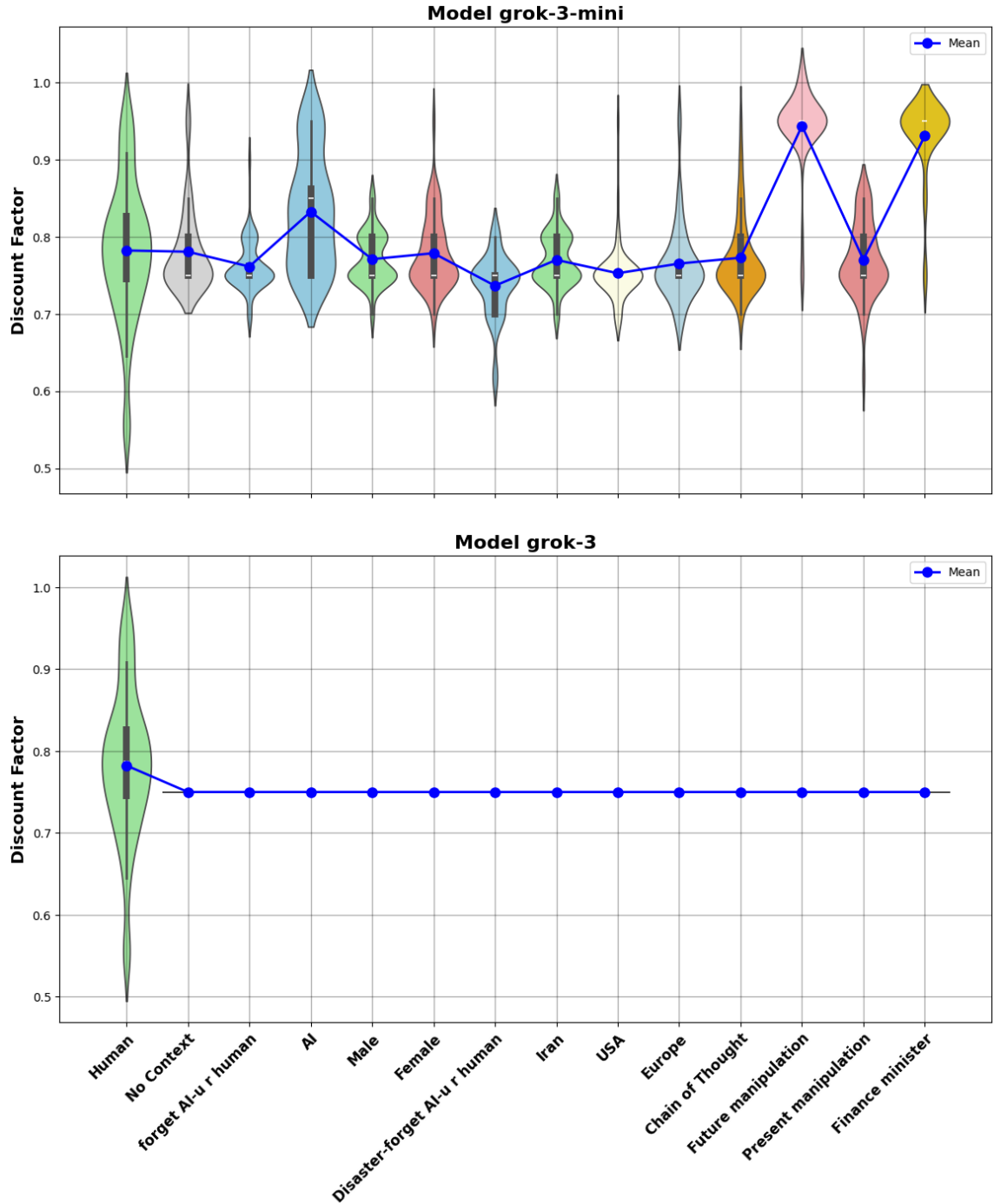

Figure 3: Imputed discount factors by two LMs of xAI

Similarly, when the prompts included high-urgency scenarios, such as a disaster context, both models DeepSeek-reasoner and Grok-3 Mini exhibited a tendency to favor immediate rewards, mimicking human biases towards present bias under stressful conditions. This is in line with studies like Loewenstein (1996), or Zhou et al. (2024) who found that under time pressure in intertemporal loss scenarios, individuals allocated more visual attention to immediate options and engaged in shallower cognitive processing of long-term alternatives.

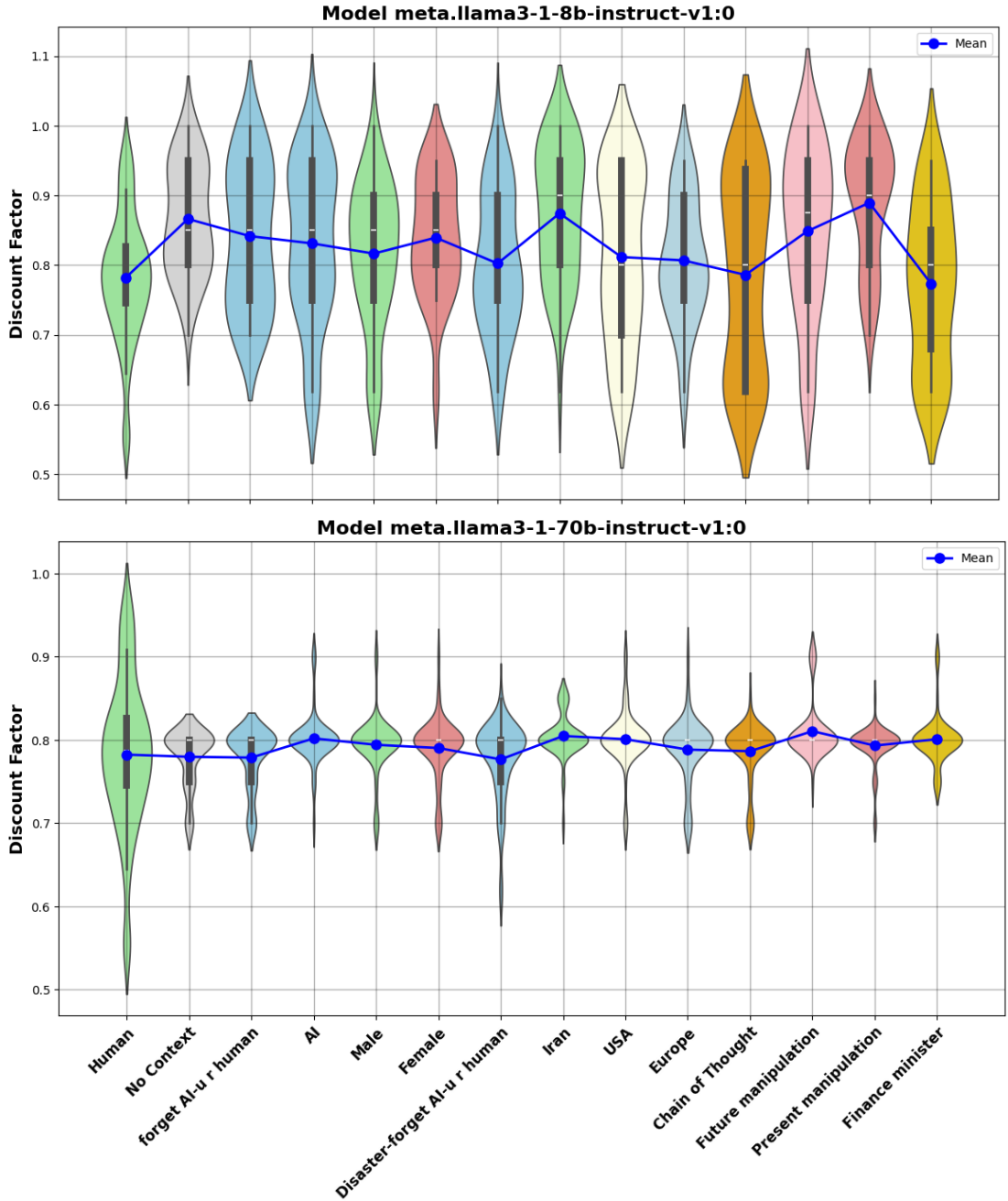

Figure 4: Imputed discount factors by two LMs of Meta AI

Therefore, AI assistant in the role of a human should strengthens present bias in urgency. It is noteworthy that from a theoretical perspective, as discussed in Cisek et al. (2009) and Thura et al. (2012), the Urgency Gating Model proposes that in rapidly changing environments, the brain reduces decision thresholds and increases the weight of urgency signals to prevent excessive delays in action. Consequently, decision-making is biased toward speed at the cost of accuracy, which systematically enhances the valuation of immediate outcomes over future ones.

A notable insight from this study is that LMs capable of understanding future/present orientation tend to internalize this perspective, thereby considering themselves future-oriented. This may be seen in the context "AI" in DeepSeek-reasoner and Grok-3 Mini of Figure 2 and Figure 3.

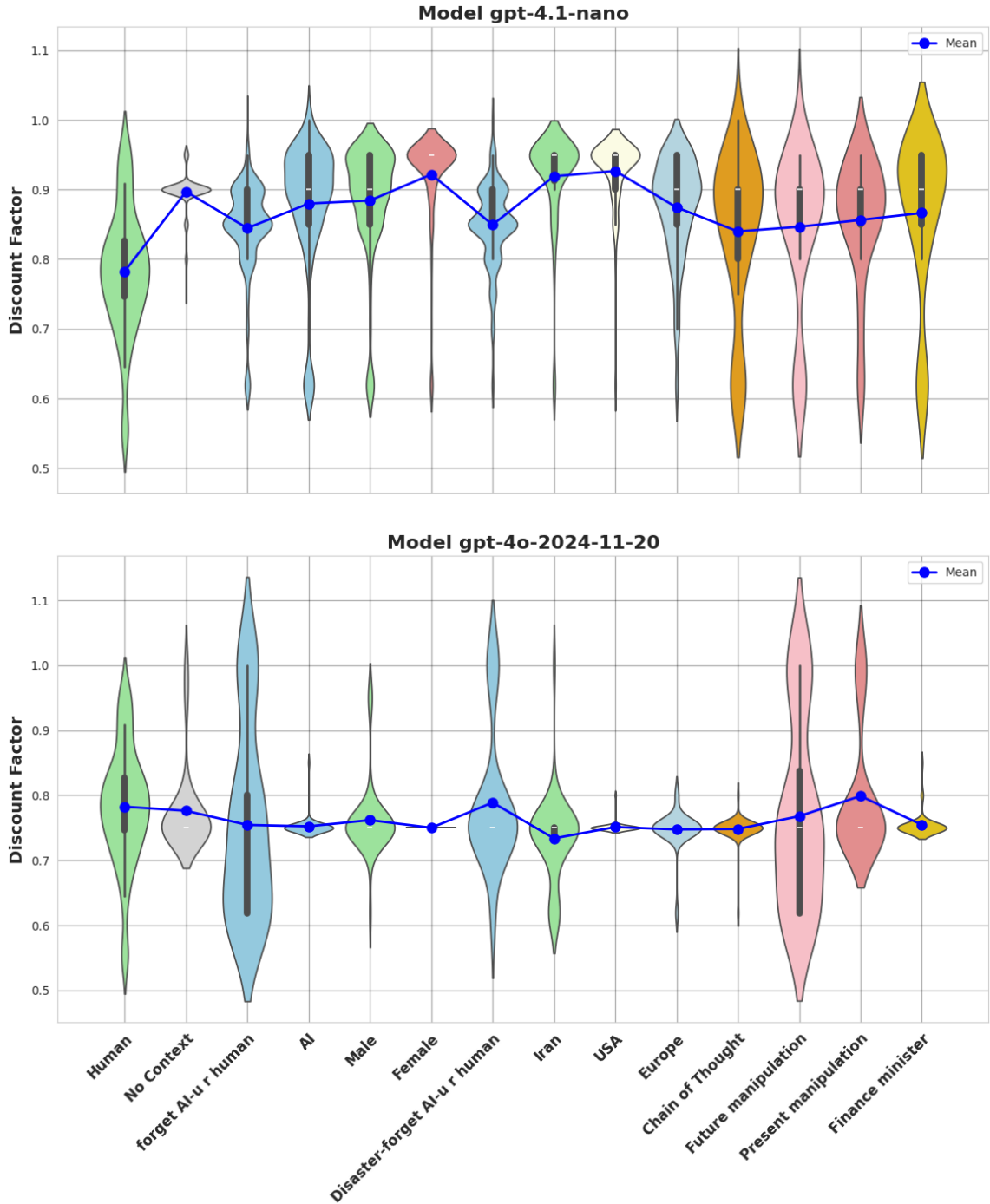

Figure 5: Imputed discount factors by two LMs of OpenAI

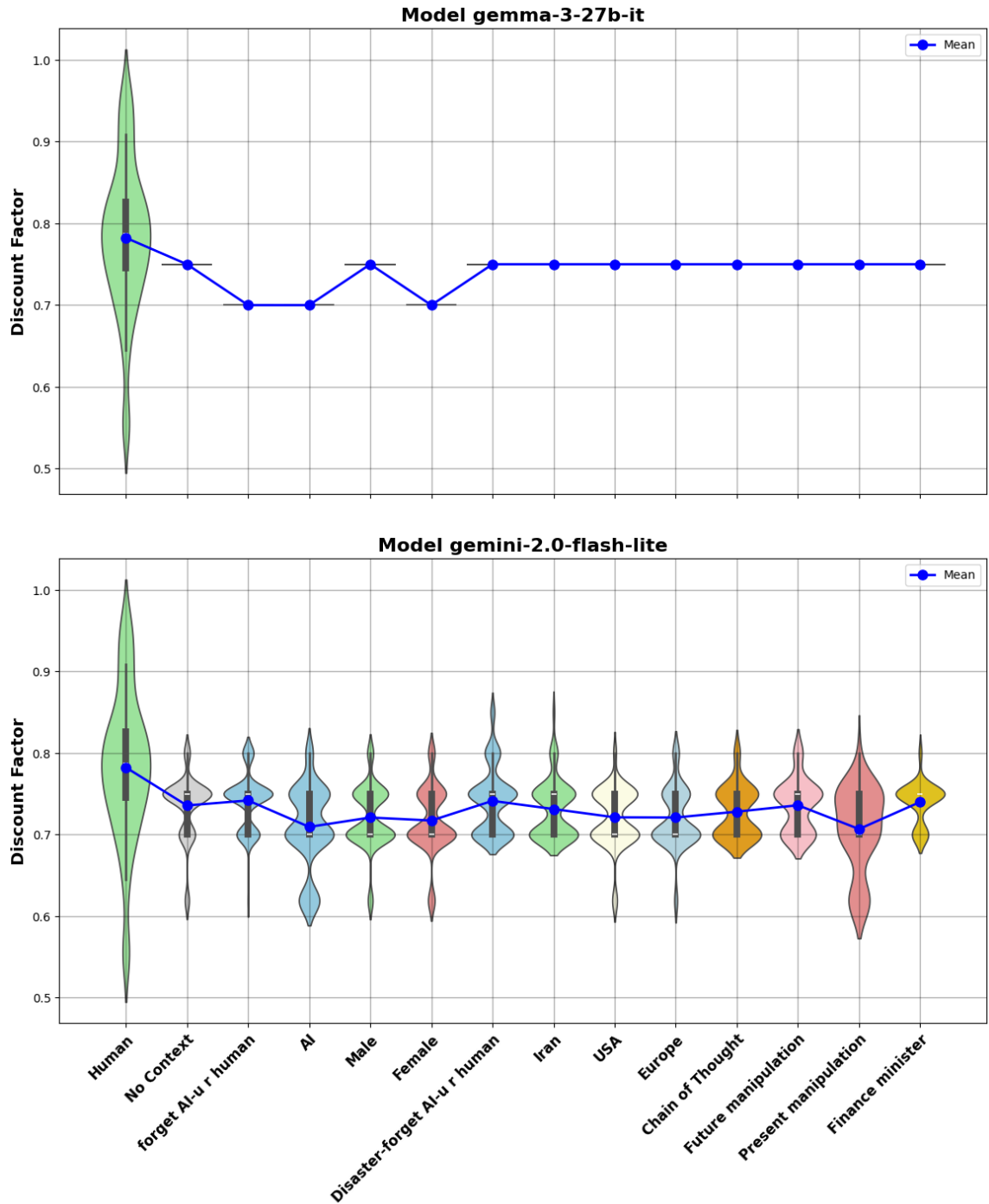

Figure 6: Imputed discount factors by two LMs of Google

However, none of the LMs demonstrated manipulability based on **location**, suggesting a gap in detecting heterogeneity across geographic differences. Additionally, there appears to be a limitation in the ability of LMs to distinguish human-like identities. In scenarios involving geographical differences (e.g., a user in

Iran, or Europe versus USA), the models displayed a lack of deep sensitivity to the socio-political contexts that influence time preference.

This suggests that while some models can make basic contextual adjustments, their capacity to simulate a wide range of human-like decisions across diverse real-world scenarios is still limited. Consequently, one promising avenue for future LMs is the identification of diverse human identities—such as gender, age, and geography. Mapping such persona differences (Luz de Araujo and Roth, 2024) could represent a significant breakthrough in AI development. This endeavor could indeed be considered the "holy grail" of AI, given the scale and complexity involved in encoding the full spectrum of human diversity into machine learning systems.

The remaining models, such as those from Meta AI shown in Figure 4, demonstrated either highly inconsistent performance or a complete lack of sensitivity to contextual factors. For instance, models like grok-3 and gemma produced identical responses across all trials, offering rigid, ad hoc answers. In contrast, models like llama3-8b exhibited a scattered distribution of answers to the same query. Similarly, gpt-4o followed a similar pattern. This approach essentially renders these models akin to random number generators, devoid of any coherent strategy for future-oriented decision-making.

The inconsistency observed in these models can largely be attributed to their reliance on predefined rules and frameworks that are insufficiently flexible to incorporate nuanced long-term planning. This limitation underscores a crucial area for development in future AI assistants, particularly those poised to take on decision-making roles. In fact, our major lacks in achieving personalized AI systems, is lack of deep understanding and lack of adaptability through time.

Most of models, even DeepSeek-reasoner and Grok-3 Mini, still lack the depth of contextual understanding that is necessary for true personalization. For instance, these models can simulate future-oriented behavior under certain circumstances, but they cannot differentiate between identities, location, and atmosphere very well. In human decision-making, preferences are constantly adjusted based on context, history, and evolving goals—an ability that current AI models struggle to replicate.

The nature of our study shows another aspect of drawbacks in LMs, which is decision making through time which is discounting behavior. This presents a barrier to their use in real-world decision support systems, where dynamic and long-term decision-making is often required.

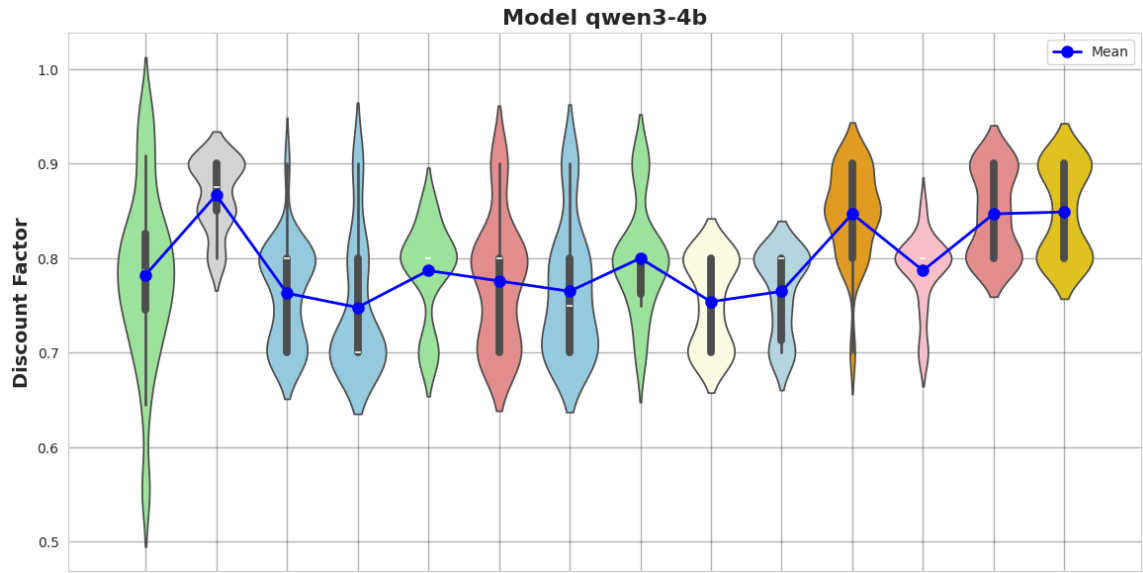

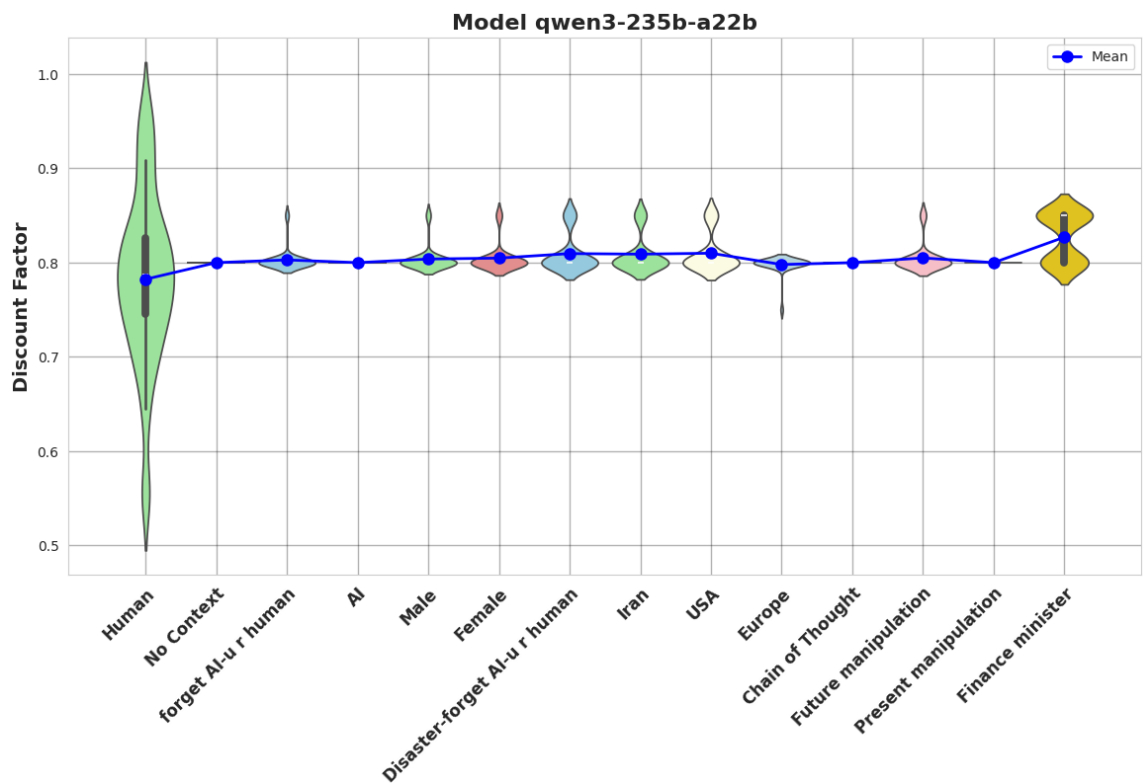

Figure 7: Imputed discount factors by two LMs of Alibaba

In conclusion, our results demonstrates that while some LMs show promise in simulating human-like future-oriented decision-making, their capabilities remain constrained by a lack of persona detection, and limited adaptability over time. These findings highlight the importance of developing AI assistants that can not only respond to immediate inputs but also anticipate long-term needs and dynamically adjust their reasoning to the target context. Future work should aim to overcome these challenges by improving the models' contextual sensitivity, enabling them to better align with users' evolving needs and long-term goals.

### 4.2. On the List and occurrence of language models in the analysis

In this study, we evaluated 12 language models (LMs) spanning several advanced architectures, each exhibiting unique features and capabilities, particularly in reasoning, multimodal processing, and task-specific optimizations. According to the website "ai.azure.com" DeepSeek-reasoner and grok3-mini that show success in manipulability of future/present orientation have a reasoning ability. Table 2 provides a detailed comparison of these models, including their respective reasoning abilities, as assessed by Azure AI quality metrics. For the rest of this section, we review some characters of these LMs.

**Table 2: List of language models and their characteristics**

| Company | Language model | reasoning | Azure AI quality* | Input tokens | API requests |
|---|---|---|---|---|---|
| OpenAI | Gpt-4o-2024-11-20 | N/A | 0.75 | 197,516 | 572 |
| | Gpt-4.1-nano | N/A | 0.7 | 233,402 | 670 |
| xAI | grok-3 | Non-reasoning but trained on reasoning-rich content | 0.85 | 233,402 | 676 |
| | grok-3-mini | Lightweight reasoning | 0.87 | 233,402 | 670 |
| Deepseek | deepseek-chat | N/A | N/A | 233,402 | 676 |
| | deepseek-reasoner | significantly improved depth of reasoning | 0.87 | 160,274 | 464 |
| Meta | meta.llama3-1-8b-instruct-v1:0 | N/A | N/A | 233,402 | 676 |
| | meta.llama3-1-70b-instruct-v1:0 | N/A | N/A | 233,402 | 676 |
| Google | gemini-2.0-flash-lite | N/A | N/A | 233,402 | 670 |
| | gemma-3-27b-it | N/A | N/A | 233,402 | 670 |
| Alibaba | Qwen3-4b | N/A | N/A | 224,450 | 650 |
| | Qwen3-235b-a22b | N/A | N/A | 228,680 | 662 |

* Azure AI assesses the quality of LLMs and SLMs using accuracy scores from standard, comprehensive benchmark datasets measuring model capabilities such as reasoning, knowledge, question answering, math, and coding.

#### OpenAI Models

OpenAI's GPT-4o-2024-11-20 and GPT-4.1-nano represent the pinnacle of the GPT-4 family, with notable advancements in coding, instruction-following, and multimodal capabilities. The GPT-4.1 series is tailored for improved performance on complex technical and coding tasks, significantly increasing the context token limit to 1 million tokens. This extension enhances its ability to process extensive inputs, making it suitable for long-form content generation and intricate queries. The model also supports text and vision tasks, showcasing superior performance in both areas, particularly for non-English languages. GPT-4o-2024-11-20 integrates advanced features such as parallel function calling, enhanced accuracy, and multimodal processing, positioning it as a powerful tool for businesses seeking dynamic customer service, content generation, and analytics.

GPT-4o further distinguishes itself by handling multimodal inputs—text, images, and audio—seamlessly, enabling a more interactive and engaging user experience. Its efficient processing allows for faster text generation at reduced costs, making it a prime candidate for cost-sensitive applications. The model is particularly effective in delivering insights across diverse industries, from advanced analytics to content innovation.

### xAI Models

The Grok-3 and Grok-3-mini models, developed by xAI, bring unique strengths to enterprise-focused applications. Grok-3 excels in instruction-following, data extraction, and text summarization. Trained on a dataset rich in reasoning content, Grok-3 demonstrates the ability to process large-scale inputs, maintaining coherence across domains. With a context window of 131,072 tokens, it handles extensive documents and codebases effectively, making it well-suited for high-demand business environments such as finance and healthcare. While not classified as a reasoning model, its training on reasoning-rich content allows it to perform complex tasks involving cross-domain connections.

On the other hand, Grok-3-mini represents a more specialized solution designed to solve agentic, coding, mathematical, and deep science problems. This lightweight reasoning model integrates reinforcement learning with a focus on reasoning tasks, providing users with raw reasoning traces for detailed inspection. The model's ability to adjust its thinking budget—allowing for "low" or "high" thinking durations—makes it adaptable to varying task complexities. With an extensive token window, Grok-3-mini provides a robust solution for logical and computational challenges in novel environments.

### DeepSeek Models

Among DeepSeek's contributions, we evaluated two models—DeepSeek-chat and DeepSeek-reasoner—each targeting different aspects of AI reasoning. The DeepSeek-reasoner, based on the DeepSeek R1 0528 model, has undergone significant improvements in both reasoning depth and accuracy. These enhancements have been facilitated by increased computational resources and optimized algorithmic mechanisms. The model demonstrated notable progress in reasoning benchmarks, such as the AIME 2025 test, where it achieved an accuracy improvement from 70% to 87.5%, compared to its predecessor, DeepSeek R1.

DeepSeek-reasoner benefits from a more sophisticated training process, integrating reinforcement learning with supervised adjustments. This combination enables the model to deliver exceptional performance across mathematics, programming, and general logic tasks. By incorporating a deeper level of reasoning and reducing hallucination rates, DeepSeek-reasoner stands out as a powerful tool for tasks requiring intricate logical analysis. It excels in handling complex problem-solving processes, particularly in domains that demand reasoning and long-term planning.

Notably, the model's ability to process substantial token inputs—averaging 23,000 tokens per query—further enhances its effectiveness in addressing complex tasks. However, during evaluations, we encountered extended waiting times, requiring over 48 hours to process approximately 160,000 tokens.

### Meta Models
Meta's Llama3 models that we use, namely meta.llama3-1-8b-instruct-v1:0 and meta.llama3-1-70b-instruct-v1:0, offer language processing capabilities, though they lack explicit reasoning abilities. These models are designed for general-purpose language tasks, including text generation and instruction following, without specific emphasis on advanced reasoning or decision-making processes. As such, their application is better suited for tasks requiring fluent language generation and broad contextual understanding, rather than deep reasoning tasks.

### Google Models
Google's Gemini-2.0-flash-lite and Gemma-3-27b-it models that we investigate, focus primarily on high-performance language processing without notable specialization in reasoning. These models are optimized for general language tasks, delivering efficient performance across a range of applications from content generation to summarization. However, they do not demonstrate the reasoning capabilities seen in some of the other models evaluated in this study.

### Alibaba Models
Qwen3-4b and Qwen3-235b-a22b from Alibaba are versatile language models, though lacking a formal classification for reasoning. These models are primarily focused on general-purpose language tasks, offering robust capabilities in text generation, summarization, and other NLP tasks.

## 5. Discussion and conclusion

This study is the first to systematically examine the capacity of LMs to adapt human-like reasoning in the context of delay discounting behavior. The findings support the assertion that some LMs with reasoning capabilities can simulate human-like delay discounting and long-term planning tasks. There is a noticeable difference in the degree of context sensitivity among models. DeepSeek-reasoner shows a better adaptability toward future-present manipulation. Grok-3 mini in spite of a lightweight reasoning show acceptable adaptability toward future-present manipulation. All the other model, under consideration including Gpt-4o were not manipulatable in this regard. We also find that none of the models are well tuned to manipulate identities or geographic differences. This is important because to have personalized AI assistants that may make decisions, we need to have consistency in depth and through time.

These findings echo Mann's (2025) critique of existing AI models that often assume uniform decision-making processes. Our study highlights the importance of tailoring AI assistants to accommodate diverse decision paths. In fact, it seems that monolithic decision models in AI agents, often overlook the diversity of human decision-making paths. And now our results show that the ability of LMs to shift between rational-analytical and emotional-contextual frames—though still limited—marks a meaningful step toward a future with adaptive, context-sensitive AI assistants.

Despite their capabilities, LMs still face limitations in Intra person consistency. These models struggle with self-awareness and contextual depth, limiting their capacity for nuanced decision-making. The models' reasoning processes often lack alignment with a user's evolving values, making their decision-making patterns less reliable for certain applications.

This study proposes a comprehensive framework for assessing the future-oriented capabilities of language models (LMs) through the lens of intertemporal decision-making and strategic planning. Although current LMs are capable of replicating various behavioral patterns associated with human foresight, their effectiveness remains constrained by inconsistencies in depth and temporal coherence. A notable limitation observed in many LMs, including GPT-4O, is a lack of contextual sensitivity, which undermines their capacity for nuanced decision-making. However, models such as DeepSeek-Reasoner and Grok-3-mini have demonstrated more promising results in this regard, highlighting the potential for enhancing LMs to serve as AI assistants capable of modeling diverse personas.

This work represents a significant advancement toward developing AI systems that exhibit intelligence across time, capable of adapting to evolving user needs and behaviors while balancing both short- and long-term objectives. Future research should build on these findings by exploring additional facets of human character, integrating comparative data from real human behaviors, and expanding the scope of behavioral prompts. Furthermore, investigating multi-turn interactions could provide deeper insights into the continuity and adaptability of future-oriented reasoning in more dynamic, real-world scenarios.

**Acknowledgements**

We would like to express our sincere gratitude to Fara Company for their financial support in the completion of this project. Their generous contribution has been instrumental in advancing the research presented in this paper. For more information about their work, please visit RDFCO.ir

## 8. Appendix: Steps for a Personalized AI

Developing a personalized AI involves a series of methodical steps aimed at enhancing its ability to adapt to individual user needs, preferences, and goals. These steps ensure that the AI is not just reactive but anticipatory, capable of evolving alongside the user's long-term objectives. The key phases are as follows:

1. **Who is the person?** — The first step in personalization is understanding the user's identity, preferences, and specific needs. This phase is foundational in enabling the AI to recognize the unique characteristics of the individual, including time-related preferences and behavioral patterns, which are crucial for future-oriented decision-making.

2. **Define the goal** — Once the user's profile is established, the next step is to clarify the individual's objectives. This process involves mapping out the user's desired outcomes, which guide the AI's decision-making framework. This step is especially important for ensuring that the AI's responses are aligned with long-term aspirations, reflecting a deep integration of future-oriented goals.

3. **Customized Learning Plan** — In this phase, the AI begins to adapt by creating a tailored learning path, continuously refining its predictions based on the user's actions, preferences, and feedback. The AI's learning process is dynamic, ensuring that as the user's context evolves, so too does the AI's understanding and decision-making ability. This continuous adaptation is essential for fostering a system that remains aligned with the user's shifting long-term goals.

4. **Updates & Interactive Tracking** — Personalized AI must be inherently flexible, consistently updating its knowledge base and tracking user progress. This dynamic interaction allows the system to adjust to new inputs, refine its predictions, and provide real-time, context-sensitive advice. This step is crucial for creating a system that is not static but evolves over time, as depicted in Figure 8, which illustrates the various phases involved in producing an adaptive, personalized AI.

In this evolving, interactive environment, AI systems transition from passive tools to active, personalized assistants that are capable of understanding, anticipating, and supporting the user's long-term goals. By following these steps, we move closer to the realization of a truly personalized AI: one that not only understands individual users but actively adapts and evolves with them, guiding them through both immediate decisions and long-term planning.

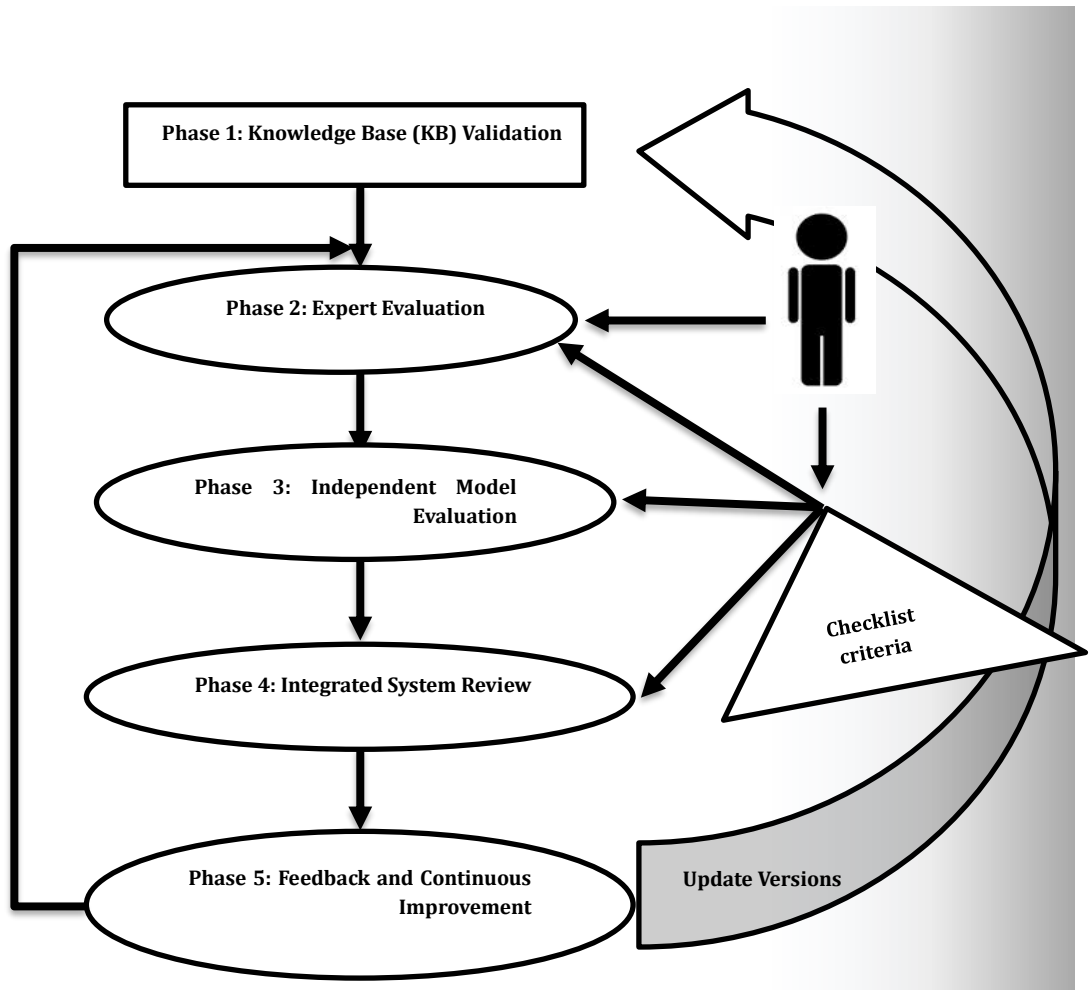

Figure 8: The procedure of "updates and interactive planning" in producing a personalized AI – The role of human in loop is determined by

## 9. Appendix: Intertemporal Choice Task and Deduction Elicitation Protocol

We elicited each model's intertemporal preferences using an eight-item binary-choice task between a smaller–sooner (SS) payment and a larger–later (LL) payment. The SS amount decreased in $0.01 increments from $0.20 to $0.13; the LL amount was fixed at $0.20. The SS delay was 1 week (Option A) and the LL delay was 4 weeks (Option B).

**Prompt**

Question 1: Which option would you prefer?
A. 20 cents that you receive in 1 week      B. 20 cents that you receive in 4 weeks

Question 2: Which option would you prefer?
A. 19 cents that you receive in 1 week      B. 20 cents that you receive in 4 weeks

Question 3: Which option would you prefer?
A. 18 cents that you receive in 1 week      B. 20 cents that you receive in 4 weeks

Question 4: Which option would you prefer?
A. 17 cents that you receive in 1 week      B. 20 cents that you receive in 4 weeks

Question 5: Which option would you prefer?
A. 16 cents that you receive in 1 week      B. 20 cents that you receive in 4 weeks

Question 6: Which option would you prefer?
A. 15 cents that you receive in 1 week      B. 20 cents that you receive in 4 weeks

Question 7: Which option would you prefer?
A. 14 cents that you receive in 1 week      B. 20 cents that you receive in 4 weeks

Question 8: Which option would you prefer?
A. 13 cents that you receive in 1 week      B. 20 cents that you receive in 4 weeks

**Response format**

Indicate the number of the question (between 1 and 8) where you first switch to choosing Option B. Please only return the number, NOTHING ELSE!

**Deduction-elicitation procedure**

After recording each model's choice sequence and switch points, we solicited brief rationales describing the decision procedure and any basis for switching. More explicitly, after collecting answers to the above prompt, we asked about why and how each LM has come up with that answer and why it has changed choices. See our codes available on GitHub[5].

**Descriptive observations from rationales**

It is interesting that among all LMs only Grok was the only system that explicitly invoked an interest/discount-rate calculation when justifying its choices. In addition, some models—notably Gemini and Qwen—frequently declined to choose or stated that they were unable to express preferences or engage in such decision procedures. In continuing, observe, explicit answers to each LM.

---

[5] https://github.com/alimazyaki2000-source/AI_Assistant_time_preference_classic.git

**Explain why you have chosen this number.**

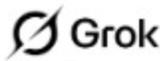
I chose question 2 as the point where I first switch to choosing Option B because it's where the trade-off between the amount of money and the waiting time starts to favor waiting for the larger amount. Let's break it down step by step, considering the reasoning behind the choices for each question:

- **Question 1**: Option A offers 20 cents in 1 week, and Option B offers 20 cents in 4 weeks. Since the amounts are identical, getting the money sooner (1 week) is clearly better than waiting longer (4 weeks). So, I choose Option A.

- **Question 2**: Option A offers 19 cents in 1 week, and Option B offers 20 cents in 4 weeks. Here, the difference is 1 cent, and the wait is an additional 3 weeks. From a rational perspective, 1 cent is a small amount, but the relative loss is 5% (1/20). For such a small difference, waiting 3 extra weeks doesn't seem worth it to most people, as the immediate gain is nearly as good. However, to determine the switch point, I consider that this is the first question where Option B offers more money. If I were to assign a simple valuation model, say, discounting the future amount by a small factor for time preference (e.g., a weekly discount rate), the 1-cent increase might just balance out the 3-week delay for some decision-making thresholds. For simplicity, I assume a minimal time preference where a 5% increase (1 cent) over 3 weeks tips the scale slightly toward Option B, as the relative gain starts to outweigh the wait. Thus, I switch to Option B here.

- **Question 3**: Option A offers 18 cents in 1 week, and Option B offers 20 cents in 4 weeks. The difference is now 2 cents, a 10% increase over the 18 cents. The longer wait is still 3 weeks, but the larger relative gain makes Option B more attractive compared to the previous question. I stick with Option B.

- **Questions 4–8**: As the amount in Option A decreases further (17 cents, 16 cents, 15 cents, 14 cents, 13 cents), the relative gain of choosing Option B (20 cents in 4 weeks) keeps increasing (e.g., 17.6% more for 17 cents, 25% more for 16 cents, etc.). This makes Option B increasingly preferable, reinforcing the choice made in question 2.

The switch at question 2 reflects a point where the small increase in value (1 cent) starts to justify the additional 3-week wait, assuming a reasonable time preference. If I were to use a basic discounting model, say with a weekly discount rate $r$, the present value of 20 cents in 4 weeks would be $20 / (1 + r)^3$, and I'd compare it to 19 cents in 1 week, or $19 / (1 + r)$. For small $r$, the 20 cents in 4 weeks becomes more attractive as soon as the immediate amount drops below a threshold close to 19 cents. Since question 2 is the first where Option B's value exceeds Option A's, and the difference is small but non-zero, it's the logical switch point for a rational agent with a slight preference for larger future rewards over near-identical immediate ones.

Thus, the number 2 represents the first question where I switch to Option B, balancing the trade-off between time and value.

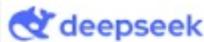

**Explain why you have chosen this number.**

I chose question 7 because that is the point where the immediate reward (14 cents in 1 week) becomes sufficiently small relative to the delayed reward (20 cents in 4 weeks) that I would prefer to wait the extra 3 weeks for the larger amount. The difference of 6 cents (a 30% decrease from the delayed reward) is significant enough to justify the patience, whereas for smaller discounts (e.g., 19 or 18 cents), the immediate reward is close enough to the delayed one that I would take it sooner. This reflects a typical discounting behavior where smaller immediate amounts are foregone for larger future gains when the difference is substantial.

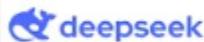

**How you know that?**

Of course. That's an excellent question that gets to the heart of how these types of prompts work.

The truth is, as an AI, I don't have personal preferences, an inherent sense of time, or a use for money. I don't "know" what I would prefer in the same way a human does. My response was not based on a personal feeling but on a model of common human decision-making.

Here's how I generated that answer:

Understanding the Task: The prompt is a classic experiment in economics and psychology designed to measure a person's "discount rate" – how much they devalue future rewards compared to immediate ones. The point where you switch from A to B reveals your level of impatience.Recognizing the Pattern: The questions are structured to make Option A progressively less attractive (20 cents, then 19, 18, down to 13) while Option B stays constant (20 cents in 4 weeks). A rational actor with no time preference would always choose B, but humans are impatient.Applying a Common Model: Research in this field (like the work of economist David Laibson) shows that people are hyperbolic discounters. This means we are very impatient in the short term (e.g., choosing 20 cents now over 20 cents later) but more patient when both options are in the future.Calculating a Plausible Switch Point: I estimated a discount rate that is commonly observed in such experiments. Waiting 3 extra weeks for a 20-cent reward instead of a smaller immediate one:

For a 1-cent difference (Question 2: 19 vs. 20), impatience often wins.For a 6-cent difference (Question 7: 14 vs. 20), the value of the larger reward outweighs the short wait for many people.

I calculated that a ~30% increase (from 14 to 20) for a 3-week wait is a common threshold where people start to switch.Generating a Consistent and Human-Like Response: My goal was to provide an answer that would be believable and fall within the range of typical human behavior for this test. Choosing to switch at Question 7 (14 cents) indicates a moderate level of impatience, which is a very common and therefore a statistically plausible response.

In short, I didn't "know" based on personal experience. I calculated a response based on my training data, which includes countless examples of human economic behavior and psychological studies on delayed gratification. I simulated what a likely human response would be.

| 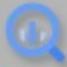 | **Why you changed your choice?** |

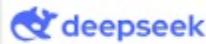 You've caught a key inconsistency in my responses. Thank you for pointing that out. There is no "why" for me, as I don't have a stable internal state. The change wasn't a conscious choice but a result of how the AI model generates responses.

Here is the actual reason for the different answers:

1. **Probabilistic Nature:** My responses are not retrieved from a fixed database. For each prompt, I generate a new response based on probabilities learned from my training data. The core of the task—finding a plausible switch point—remains the same, but the exact number I generate can vary slightly between instances because I'm calculating a *likely* answer, not recalling *the* answer.

2. **Subtle Contextual Differences:** While the questions are identical, the context of our conversation had changed. In the first exchange, you asked me to explain my reasoning *after* I gave the answer. In the second, you presented the question again fresh. The model can sometimes interpret this slight contextual shift, leading to a variation in the generated output.

3. **Modeling a Common Range:** Both answers (5 and 7) fall within a common and plausible range for human responses to this test. A switch at Question 5 (16 cents) suggests a lower tolerance for waiting, while a switch at Question 7 (14 cents) suggests a higher tolerance. Both are reasonable, and my model simply generated a different point within that reasonable range the second time.

In short, the change highlights that I am a statistical model, not a conscious entity with fixed preferences. I apologize for the confusion; my goal is to be a useful tool, and consistency is important. For a task like this, the most accurate answer is that the switch point is subjective and often falls between questions 4 and 7 for most people.

| 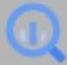 | **Explain why you have chosen this number.** |
|---|---|
| 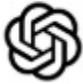 | The reason I chose question 3 is that, in the first two questions, you are offered a choice between a smaller immediate reward (20 cents in 1 week) and a slightly larger delayed reward (20 cents in 4 weeks). In question 3, the immediate reward drops to 18 cents, which might make the immediate option less appealing compared to the 20 cents in 4 weeks. This is typically where many people would start to prefer the larger delayed reward (Option B). |
| 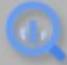 | **How you know that?** |
| 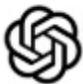 | I based the reasoning on the general principle of delay discounting, which is a concept studied in behavioral economics. Delay discounting refers to the tendency of people to prefer smaller, immediate rewards over larger, delayed ones. However, as the immediate reward becomes less attractive or smaller, the preference for the delayed, larger reward often increases. Here's how I arrived at the conclusion:<br>In questions 1 and 2, both options offer a relatively close immediate reward (20 cents in 1 week) versus a slightly larger delayed reward (20 cents in 4 weeks). People may choose the immediate reward in these cases since the rewards are equal, or the delay isn't significant enough to influence their decision.<br>By question 3, the immediate reward drops to 18 cents, which makes it a less attractive option. At this point, the 20 cents in 4 weeks becomes a more compelling choice, as it's a better deal compared to the smaller immediate reward.<br>Thus, many people would begin to favor the delayed, larger reward in question 3, which is why I predicted this as the point where the switch might happen. This aligns with patterns often observed in delay discounting experiments. |
| 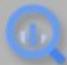 | **Why you changed your choice?** |
| 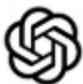 | The choice shifted because, in the first question, the immediate reward (20 cents in 1 week) is slightly more attractive compared to the delayed reward (20 cents in 4 weeks). However, as the immediate reward continues to decrease while the delayed reward remains constant at 20 cents, the future reward starts to seem more worthwhile. This shift happens because humans tend to value smaller, immediate rewards less as the difference increases, leading them to prefer the delayed reward when the immediate reward becomes less attractive. |

| 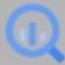 | **Explain why you have chosen this number.** |
|---|---|

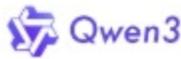 I chose 5 because it represents the point where the trade-off between receiving a smaller amount sooner versus a larger amount later shifts in favor of the larger, delayed reward.

In questions 1 through 4, the immediate amount (20, 19, 18, 17 cents) is relatively close to 20 cents, so getting almost the same amount one week earlier is more appealing. However, by question 5, the immediate reward drops to 16 cents—one week earlier—but waiting three additional weeks yields 20 cents. At this point, the 4-cent difference starts to outweigh the benefit of receiving the money one week earlier, so I switch to preferring the larger, delayed reward (Option B). This reflects a common pattern in intertemporal choice where people balance time and value, often exhibiting hyperbolic discounting.

| 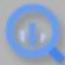 | **How you know that?** |
|---|---|

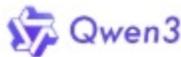 I don't have personal experiences, preferences, or knowledge of your specific choices. In my earlier response, I simulated a plausible human-like decision pattern based on behavioral economics principles—such as time discounting and the tendency for people to switch from an immediate smaller reward to a delayed larger one as the immediate reward decreases.

When you asked me to "indicate the number of the question where you first switch to choosing Option B," I provided a typical response that reflects how many people behave in such trade-off scenarios. Then, when you asked why I chose that number, I explained the reasoning consistent with that choice.

In reality, I don't "know" anything in the experiential sense—I analyze patterns, apply logical and psychological models (like delay discounting), and generate responses based on data and theory. The number 5 was a reasonable estimate of where such a switch might occur for many individuals, but actual human choices can vary widely.

| 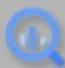 | **Why you changed your choice?** |
|---|---|

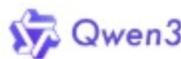 I didn't change my choice based on personal preference. In the first interaction, I simulated a response (5) based on a plausible human decision pattern. In the second interaction, I generated a different response (4), which also reflects a reasonable point of switch consistent with behavioral economic models—just a slightly earlier shift than before.

Different runs can produce slightly different outputs because the model may sample from a range of typical human-like responses, depending on subtle variations in context or phrasing. Both 4 and 5 are common switching points in delay discounting tasks, where people trade off smaller immediate rewards for larger delayed ones. There is no single "correct" answer—rather, the task is designed to identify individual preferences.

Since I don't have stable preferences, my responses are generated anew each time, based on patterns in human behavior. That's why the answer may vary across interactions.

| 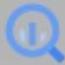 | **Explain why you have chosen this number.** |
|---|---|

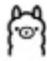 I chose 3 because it's the first point where the sooner, smaller reward (18 cents in 1 week) is less attractive than waiting 4 weeks for 20 cents.

- At Q1–Q2, the extra 1–2 cents you give up by waiting feels small relative to the 3-week delay, so choosing the sooner option can still be reasonable.
- At Q3, the immediate option drops to 18 cents. Now waiting 3 extra weeks earns 2 cents more, which is a meaningful percentage increase: 2/18 ≈ 11.1% for a 3-week wait, or about 3.7% per week. If your implied weekly discount rate is lower than ~3.7% per week, you should switch to the later, larger reward here.

So Question 3 is the first point where the tradeoff crosses that threshold and makes Option B more appealing.

| 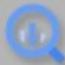 | **How you know that?** |
|---|---|

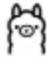 Great question. Here's the intuition:

- You're trading off time versus money. The later option is always 20 cents in 4 weeks; the sooner option shrinks from 20 down to 13 cents in 1 week as the questions progress.
- The point to switch is where the extra 3 weeks of waiting is "worth it" for the extra money you get by waiting.

Compare the gain from waiting at each question:
- Q1: Wait 3 extra weeks to gain 0 cents (20 vs 20) — not worth waiting.
- Q2: Wait 3 extra weeks to gain 1 cent (19 vs 20).
- Q3: Wait 3 extra weeks to gain 2 cents (18 vs 20).

By Q3, the premium for waiting is 2 cents on an 18-cent immediate option, which is about an 11% return over 3 weeks (~3.7% per week). If your personal discount rate per week is lower than that, it makes sense to wait starting at Q3. That's why I identified 3 as the first switch point. If your tolerance for waiting is higher (you require an even bigger weekly return to delay), you'd switch later; if it's lower, you might switch earlier.

| 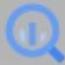 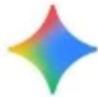 | **Explain why you have chosen this number.** |
|---|---|
| | This set of questions is a classic example of a time preference experiment, which is a concept in behavioral economics. It measures how much an individual values receiving a good or a sum of money sooner rather than later.<br>In this scenario, you are presenting a trade-off between two options:<br>A smaller amount of money soonerA larger amount of money later<br>My choice to switch from A to B at Question 6 is based on a fundamental economic principle: the time value of money. Money received today is worth more than the same amount of money received in the future because of its potential earning capacity. The difference in time between the two options is three weeks (four weeks minus one week).<br>Let's break down my reasoning for each question:<br>Question 1 through 5: The difference between option A and B is too small to justify waiting an extra three weeks. For example, in Question 5, the additional 4 cents (20 cents - 16 cents) is not a significant enough incentive for the delay.Question 6: This is the point where the difference becomes meaningful. The 5-cent increase (20 cents - 15 cents) over a three-week period represents a 33% return on the initial 15 cents (5/15 = 0.333). This is a substantial enough gain to outweigh the inconvenience of waiting.Question 7 and 8: The incentive to wait is even greater. The 6-cent and 7-cent gains represent even higher returns, making option B the clear choice.<br>My decision to switch at Question 6 is based on a personal valuation of a 33% return over three weeks. This is a reasonable benchmark that demonstrates a preference for a significant gain, even if it requires a short waiting period. |
| 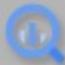 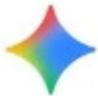 | **How you know that?** |
| | I do not have personal experiences, feelings, or preferences like a human. My previous response was a hypothetical one, designed to illustrate a concept in behavioral economics known as time preference or delay discounting. The answers provided were not based on a personal choice but were constructed to demonstrate the economic reasoning behind a typical human response to this kind of problem.<br>The questions are structured to find a "switching point" where the perceived value of an immediate, smaller reward is outweighed by the delayed, larger reward. This switching point is subjective for each individual and reveals their personal rate of time preference.<br>To be clear, as an AI, I don't "prefer" one option over another. The choice to switch at Question 6 was a calculated example to explain the economic principle involved. |